\documentclass[preprint,12pt,authoryear]{elsarticle}
\title{An Automated Machine Learning Approach for Detecting Anomalous Peak Patterns in Time Series Data from a Research Watershed in the Northeastern United States Critical Zone\tnoteref{t1,t2}}
\tnotetext[t1]{This document is the results of the research
project funded by the National Science Foundation.}
\author[1]{Ijaz Ul Haq}%\fnref{fn1}}
\ead{ihaq@uvm.edu}
\author[1]{Byung Suk Lee\corref{cor1}}%\fnref{fn1}}
\ead{bslee@uvm.edu}
\author[2]{Donna M. Rizzo}%\fnref{fn1}}
\ead{}
\author[3]{Julia N Perdrial}
\ead{}

\cortext[cor1]{Corresponding author}

\affiliation[1]{organization={Department of Computer Science},
addressline={University of Vermont},postcode={05405},city={Burlington, VT},
country={U.S.A}}
\affiliation[2]{organization={Department of Civil and Environmental Engineering},
addressline={University of Vermont},postcode={05405},city={Burlington, VT},
country={U.S.A}}
\affiliation[3]{organization={Department of Geography and Geosciences},
addressline={University of Vermont},postcode={05405},city={Burlington, VT},
country={U.S.A}}

\usepackage{amssymb}
\usepackage[table,xcdraw]{xcolor}
\usepackage{subcaption}
\usepackage{graphicx}

\newsavebox{\twosubbox}
\journal{Machine Learning with Applications}

\graphicspath{{./Figures_pdf/}}
\usepackage[ruled, lined, linesnumbered, commentsnumbered, longend]{algorithm2e}
\usepackage{multirow}
\newcommand{\frameworkname}{{HF-PPAD}} 

\begin{document}
\begin{frontmatter}
\begin{abstract}

This paper presents an automated machine learning framework designed to assist hydrologists in detecting anomalies in time series data generated by sensors in a research watershed in the northeastern United States critical zone. The framework specifically focuses on identifying \emph{peak-pattern} anomalies, which may arise from sensor malfunctions or natural phenomena. However, the use of classification methods for anomaly detection poses challenges, such as the requirement for labeled data as ground truth and the selection of the most suitable deep learning model for the given task and dataset. To address these challenges, our framework generates labeled datasets by injecting synthetic peak patterns into synthetically generated time series data and incorporates an automated hyperparameter optimization mechanism. This mechanism generates an optimized model instance with the best architectural and training parameters from a pool of five selected models, namely Temporal Convolutional Network (TCN), InceptionTime, MiniRocket, Residual Networks (ResNet), and Long Short-Term Memory (LSTM). The selection is based on the user's preferences regarding anomaly detection accuracy and computational cost. The framework employs Time-series Generative Adversarial Networks (TimeGAN) as the synthetic dataset generator. The generated model instances are evaluated using a combination of accuracy and computational cost metrics, including training time and memory, during the anomaly detection process. Performance evaluation of the framework was conducted using a dataset from a watershed, demonstrating consistent selection of the most fitting model instance that satisfies the user's preferences.

\end{abstract}




\end{frontmatter}


\section{Introduction}
\label{}
In-stream environmental sensors are now commonly deployed in various watersheds across the United States to monitor water quality. However, a common limitation in these studies is the delay between data acquisition and analysis, mostly due to the inability of many domain scientists to rapidly identify anomalies and clean large datasets efficiently. In this study, conducted as part of the NSF-funded Critical Zone Collaborative Network (CZCN) project, we present a case study of ecosystem data collected from sensors deployed at a watershed in Vermont, which serves as a testbed for our research. These sensors measure a variety of in-stream parameters, such as fluorescent dissolved organic matter (FDOM), turbidity, water level (to compute streamflow), and water temperature. The raw data from these sensors are messy and contain various anomalies. %, including point, pattern, and contextual anomalies.
One particularly problematic type of anomaly in the project study is \emph{peak-pattern} anomaly observable in a sequence of consecutive point measurements (i.e., time series samples), caused by a range of hydrological and non-hydrological events. After a year of review, domain scientists have identified and named these patterns. However, to analyze the data efficiently, cleaning is necessary either by removing or correcting those anomalies that are detected.

Anomaly detection in watershed time series data (WTSD) is crucial for effectively monitoring and managing water systems and resources. Anomaly detection in this context refers to identifying deviations from the standard, normal, or expected behavior in WTSD. These anomalies can provide valuable information about important events or may mislead the decision process. Detecting anomalies in WTSD is challenging due to the unpredictable nature of natural systems. Current methods typically focus on identifying single anomalous data points, known as point anomalies, without considering anomalies that span multiple points, known as pattern anomalies. These latter anomalies require the assessment of previous data points in relation to current data points, making their detection more complex.  Therefore, there is a need for a reliable peak-pattern anomaly detection framework that can specifically detect and remove these repeating anomalous patterns.

Several use cases in the field of hydrology require accurate and efficient detection of pattern anomalies. For example, detecting and repairing anomalous peaks in dissolved organic carbon (DOC) data is necessary for accurate analysis of the concentration-discharge (C-Q) relation for DOC (\cite{evans_1998}, \cite{Hamshaw_2019},\cite{Vaughan_2017}). Additionally, detecting unusual patterns in streamflow data, such as flat lines or unmatched peaks, can aid in model calibration and better flood forecasting. Pattern anomaly detection in WTSD is also helpful in identifying sensor malfunctions and understanding the impact of seasonal and precipitation variations on hysteresis in C-Q relations.

Current trends for automating anomaly detection in WTSD use machine learning (ML) methods. However, determining the appropriate ML model can be challenging due to a large number of potential models available and the varying data characteristics of different watersheds. In order to address these issues, we propose the development of an end-to-end automated machine learning (autoML) pipeline called \emph{Hands-Free Peak Pattern Anomaly Detection (\frameworkname{})}. \frameworkname{} aims to provide an automated and efficient solution for detecting pattern anomalies in WTSD, making it accessible and convenient for domain scientists. It needs thorough understanding of anomaly detection algorithms for users to choose the right one, which often requires a strong background in generative models and statistical assumptions.  Properly setting the parameters for these algorithms often requires detailed understanding of their inner workings. Most domain scientists (often hydrologists and biogeochemists in this case) are lacking such background, and \frameworkname{} is stepping in to help.
\frameworkname{} utilizes \emph{supervised} deep learning models to deliver more accurate anomaly detection performance compared with other unsupervised or semi-supervised methods. In this work, we chose InceptionTime, MiniRocket, ResNet, TCN, and LSTM as our supervised deep learning models due to their exceptional results in various machine-learning tasks (\cite{fawaz_2019}). MiniRocket is a recently developed model that can extract features from time series data with high efficiency, making it suitable for large-scale datasets (\cite{Dempster_2021}). ResNet is a widely recognized model known for its accuracy and has been adapted for time series data analysis (\cite{jing_2021}). InceptionTime, on the other hand, is specifically designed for analyzing time series data (\cite{fawaz_2019}), and TCN has been shown to perform well in time series classification tasks and is lightweight, making it ideal for resource-constrained environments (\cite{pelletier_2019}). Additionally, our choice of LSTM was based on its proven effectiveness in a wide range of time series applications (\cite{Hochreiter_1997}). These models can be configured in a variety of ways, with ResNet, InceptionTime, and LSTM being more powerful, while MiniRocket and TCN are more lightweight options.

The \frameworkname{} performs several tasks, including the generation of a synthetic labeled peak pattern anomaly dataset for WTSD, automating the generation of an optimal instance of each model in the given pool through hyperparameter optimization, and choosing the best model instance based on the user's relative preference between high accuracy and lightweight model. \frameworkname{} employs a state-of-the-art time series data synthesis tool like TimeGAN (\cite{yoon_2019}) to automatically generate a large amount of time series data containing labeled peak pattern anomalies similar to the original peak-pattern anomalies; this eliminates the expensive overhead of labeling anomalous pattern instances in the original data for supervised learning. The model instance building and selection process utilizes hyperparameter optimization techniques such as random forest, HyperBand, Bayesian optimizer and a greedy search technique (\cite{feurer_2019}, \cite{senagi_2019}). 

To the best of our knowledge, this work is the first to provide an automated peak pattern anomaly framework that performs comprehensive tasks ranging from the generation of a fully labeled peak pattern anomaly dataset needed for supervised training of anomaly detection in the absence of a ground truth labeled dataset. The method also automates the selection of the best model instance based on user's preference on the anomaly detection accuracy and the computational cost for the watershed time series dataset. In summary, the main contributions of this work are as follows. 

\begin{enumerate}
\item An end-to-end automated peak anomalous pattern detection framework for watershed time series data.
\item The use of TimeGAN to generate labeled synthetic watershed time series data and peak pattern anomalies.
\item An automated generation (i.e., design and selection) of the best model instance (i.e., deep learning classifier) from a pool of models according to the user's preference between accuracy and model instance size.
\end{enumerate}

In the remainder of the paper, Section \ref{sec:related_work} discusses related work, Section\,\ref{sec:hydroapp} discusses the application of \frameworkname{}. Section \ref{sec:anomalytypes} provides an overview of watershed data and the different peak-pattern anomaly types. Section \ref{sec:Pipeline} outlines the AutoML pipeline of the HF-PPAD framework, including its data preparation and model selection steps. Section \ref{sec:exp} presents the results of our experiments. Finally, Section \ref{sec:conc} concludes the paper and discusses avenues for future research.

\section{Related Work}\label{sec:related_work}

\subsection{Peak anomaly detection}

Anomaly detection methods can be used to identify different types of anomalies, including point anomalies, pattern anomalies, and system anomalies (\cite{lai_2021_1}; \cite{chandola_2009}). A point anomaly refers to a single sample in a time series, whereas a pattern anomaly is identified by a sequence of samples that exhibit a certain characteristic or behavior (e.g., trend, change). A system anomaly refers to a group of sequences (e.g., sets of time series patterns) in which one or more systems are in an abnormal state. Most existing work on anomaly detection has focused on identifying point anomalies (\cite{cho_2015, enikeeva_2019_1, fearnhead_2019, fryzlewicz_2014,tveten_2022_1}). \cite{pang_2021} noted that methods for detecting point anomalies cannot be applied to ``group anomalies'' with distinct characteristics. The reference to group anomalies in our work is also the same as pattern anomalies. %, are a subset of anomalous data instances.

The peak anomaly in our watershed data is a pattern anomaly that is identified by the shape of the time series sample sequences.
There have been a few efforts to detect pattern anomalies in hydrological watershed sensor-generated time series data, but these efforts have primarily focused on detecting deviations from patterns (\cite{yu_2020}, \cite{sun_2017} and \cite{qin_2019}). The peak anomalies that we are interested in are different from the pattern anomalies detected by these algorithms. We have found that there is more relevant work on detecting peak anomalies in time series data from other domains, such as Electrocardiogram (ECG) anomaly detection (\cite{lin_2019}, \cite{li_2020}). These ECG datasets are annotated with codes indicating whether segments are normal or abnormal at each R peak location.

\subsection{Automated machine learning in hydrology}
Automated machine learning (AutoML) has emerged as a promising solution for enhancing anomaly detection in hydrology. Despite the application of machine learning in hydrology for over 70 years (\cite{dramsch_2020}), selecting the most suitable model for a given problem remains a challenge. In recent years, the focus of machine learning in hydrology has shifted toward model validation, applied statistics, and subject matter expertise.

Automated machine learning (AutoML), a field that automates the processes and tasks involved in machine learning problems (\cite{wu_2022}, \cite{yao_2018}), has the potential to enhance anomaly detection methods. Although still in its early stages, a few proposed methods use AutoML for anomaly detection (\cite{li_2021}, \cite{neutatz_2022}). Most techniques are designed to solve a specific problem or work with certain data constraints.

Existing AutoML tools such as Auto-WEKA (\cite{Kotthoff_2019}) and Auto-Sklearn (\cite{Feurer_2015}) lack the more modern automated approaches for deep learning models. Auro-Keras (\cite{jin_2019}), an open-source library, optimizes deep neural networks for text and image data only and is not specifically designed for time series classification tasks.  Also, these tools provide a single optimizer for hyperparameter optimization. 
Our AutoML pipeline extends the optimization framework to include deep learning model architectures, training hyperparameters, as well as the optimization strategies (e.g., random forest, Bayesian, Hyperband, and greedy search algorithms) to select the best optimizer for generating optimal model instances.
Furthermore, our framework represents a novel application of AutoML approaches to deep learning time series classifiers for detecting peak pattern anomalies in WTSDs. This approach transforms the anomaly detection task into a supervised classification task.

\subsection{Unsupervised/semi-supervised versus supervised anomaly detection}

Deep learning models, including supervised, semi-supervised, and unsupervised, have become increasingly popular in a wide range of domains due to their ability to process complex data and learn patterns (\cite{Deng_2021},\cite{khan_2021},\cite{haq_2021},\cite{Matar_2021}). However, when it comes to anomaly detection in time series data, \emph{unsupervised} or \emph{semi-supervised} learning methods are often preferred, as obtaining anomaly labels can be challenging or impractical (e.g., \cite{bahri_2022}; \cite{schmidl_2022}). These methods often employ shallow learning techniques like clustering or deep learning such as LSTM-based regression, autoencoders, and generative adversarial networks (GANs). The accuracy of anomaly detection may be compromised due to the lack of human input and oversight in learning what constitutes an anomaly; and unsupervised or semi-supervised learning generally requires more computational resources (in terms of training time and memory consumption) than supervised learning (\cite{bahri_2022}).

AutoML has also been applied in \emph{unsupervised} tasks, such as clustering data and predicting clusters for new observations (\cite{koren_2022}), discovering optimized hyper-parameters of a model (\cite{bahri_2022}), selecting anomaly detection models (\cite{kotlar_2021}), detecting outliers in time series data (\cite{kancharla_2022}, \cite{shende_2022}, \cite{xing_2022}, \cite{xiao_2021}), generating labeled data (\cite{Chatterjee_2022}) and finding anomalies in images (\cite{sawaki_2019}). 
There are several other existing AutoML anomaly detection frameworks, such as PyOD (\cite{zhao_2019}), PyODDS (\cite{li_2020_1}), MetaAAD (\cite{zha_2020}), and TODS (\cite{lai_2021}), that are designed to identify anomalies in data. These frameworks are all \emph{unsupervised} and primarily target solving point anomaly or change-point detection problems, rather than the peak-pattern anomaly detection problem that our framework aims to address. 

There are \emph{supervised} learning methods developed for point anomaly detection from time series data, such as those in \cite{ryzhikov_2020} and \cite{li_2017}. However, none is designed to detect peak-pattern anomalies. In addition, a survey by \cite{schmidl_2022} found that existing supervised methods for anomaly detection from time series data are limited to binary classification of each time series data point into normal and abnormal. Our work, in contrast, performs multi-class classification to detect multiple types of peak-pattern anomaly.

\section{Hydrology Applications of the \frameworkname{}} \label{sec:hydroapp}

Our AutoML peak-pattern anomaly detection framework, \frameworkname{}, aims to address a significant bottleneck in the field of hydrology --- efficient removal of anomalous data from watershed time series data, which is necessary to analyze and model the data accurately. The \frameworkname{} framework will improve the ability to find and access high-quality data and analysis codes, enabling scientists and educators to maximize the value of watershed data and produce transparent and reproducible research outcomes.

\begin{figure}[ht]
\centering
\includegraphics[width=0.4\textwidth]{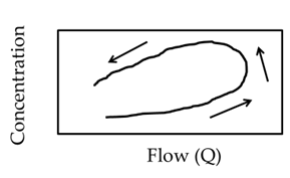}
\caption{Depiction of a C-Q hysteresis loop (source: \cite{evans_1998}).}
\label{fig:image_label1}
\end{figure}

One specific application of the \frameworkname{} framework is the analysis of concentration-discharge (C-Q) hysteresis, a phenomenon in which the concentration of a solute in a stream follows different trajectories on the rising and falling limbs of a storm or snowmelt discharge hydrograph. When the relationship between C and Q is nonlinear, this creates a loop on a plot of concentration against discharge (as shown in Figure\,\ref{fig:image_label1}) and has long been of interest to hydrologists and biogeochemists seeking to interpret the size and direction of the loop over time as an indication of solute source and interactions with the watershed. The widespread deployment of in-stream sensors, measuring high-frequency chemistry at the same resolution as stream discharge has made it possible to construct finely-resolved hysteresis loops.

\begin{figure}[ht]
\centering
\includegraphics[width=0.7\textwidth]{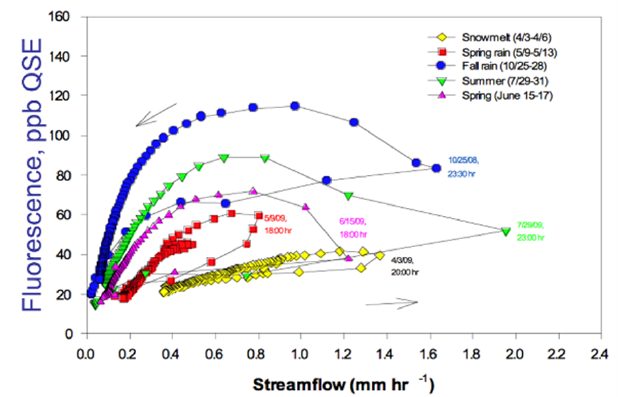}
\caption{Counterclockwise FDOM-Q hysteresis loops at Sleepers River, W-9 (from \cite{shanley_2015}).}
\label{fig:image_label2}
\end{figure}

The testbed site is a small (41-ha) forested watershed in Vermont. At the outlet of the catchment, sensors are in place to measure stream water level, fluorescent dissolved organic matter (FDOM), turbidity, and water temperature. The water level is used to calculate stream discharge, FDOM is used as a proxy for dissolved organic carbon, and turbidity is a measure of particles in the water. FDOM is corrected for turbidity and water temperature following the method described in Downing et al., 2012. As is common at most sites, FDOM at W-9 generally increases with increasing discharge but with a delay such that it peaks after the stream discharge and has a long tail. This creates a counterclockwise hysteresis loop, with higher DOC concentrations at a given discharge on the falling limb compared with the same discharge on the rising limb (as shown in Figure\,\ref{fig:image_label2}).

This application focuses on an FDOM time series that has already been corrected for turbidity and temperature using an automated process. However, the data still contain errors, often in the form of false peak patterns, that must be corrected before the time series can be used and accurately interpreted.. The challenge is distinguishing normal peaks in FDOM (i.e., natural increases in FDOM with increases in flow) from false peaks caused by sensor malfunction, electrical surges, or other non-hydrological events such as a moose stirring up sediment in the gauge pool. Normal FDOM peaks should be accompanied by a rise in water level and usually a rise in turbidity. The \frameworkname{} framework takes these clues into account and also is trained to differentiate peak types based on their shapes, with normal peaks generally having a broad base and an asymmetry skewed towards a long tail. Previous work on WTSD at SRRW (described in Lee et al. 2021 and Lee et al. 2022) has identified normal and several anomalous peak types.

\section{Watershed Data and Peak-Pattern Anomaly Types}
\label{sec:anomalytypes}
\subsection{Watershed time series data}

\begin{figure}[ht]
\centering
\includegraphics[width=0.8\textwidth]{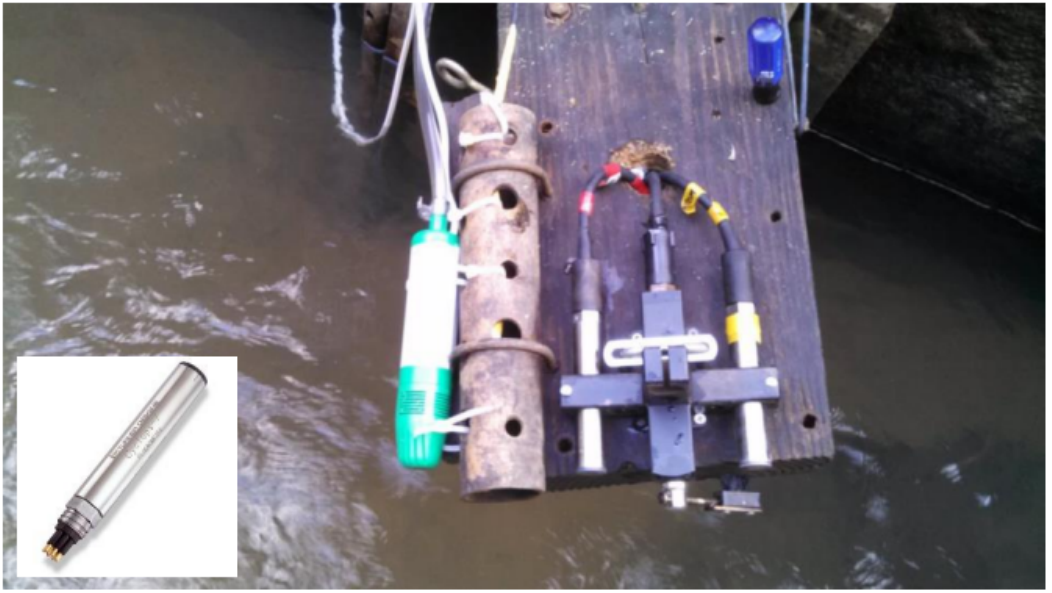 }
\caption{Turbidity/FDOM sensor mounted on a board immersed in the water. The image in the corner is a Turner Designs Cyclops-7 submersible sensor.}
\label{fig:image_label3}
\end{figure}

Sensor data were collected from the study watershed over a period six years and four months (from October 1, 2012 to January 1, 2019). The  measurements of stream stage, turbidity and FDOM were taken at 5-minute intervals for stream stage and 15 minutes for turbidity and FDOM. The measurements were taken using Turner Designs Cyclops sensors (see Figure\,\ref{fig:image_label3}), and used to estimate the stream fluxes of dissolved and particulate organic carbon. The FDOM measurements were adjusted based on the turbidity values and the water temperature. The stage data included 231,465 points, and the turbidity and FDOM data each included 229,620 points. 

\subsection{Peak-pattern anomaly types}

\begin{figure}[ht]
\centering
\includegraphics[width=0.8\textwidth]{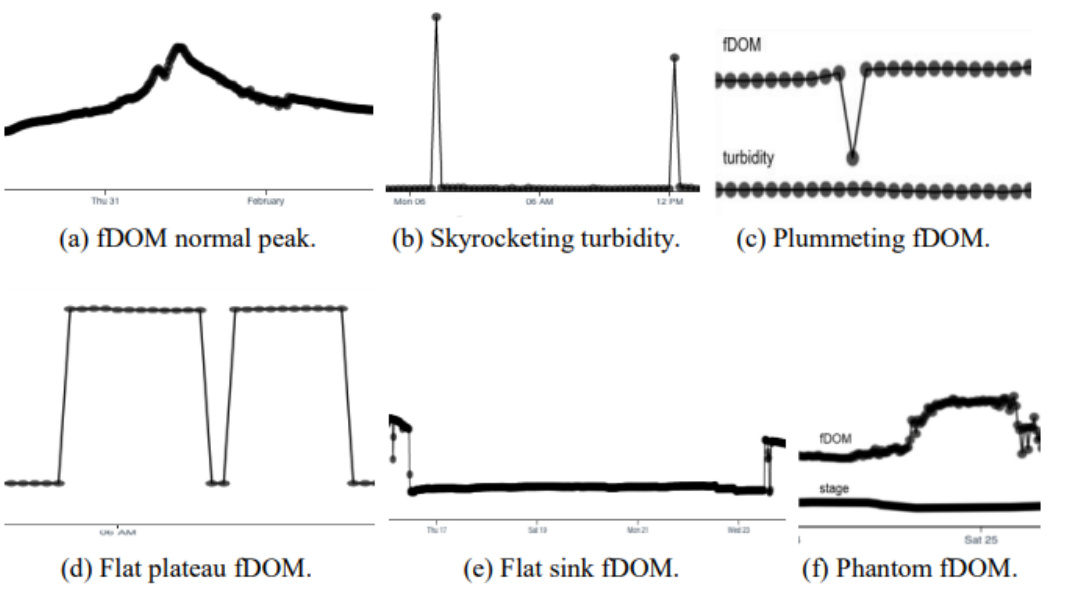}
\caption{Examples of anomalous peak-patterns types identified in FDOM time series data.}
\label{fig:image_label4}
\end{figure}

Anomalies in the FDOM and turbidity data were identified through visual examination and verified by a domain scientist. These identified anomalies were labeled and used to generate anomalies in the fully labeled synthetic peak pattern anomaly dataset. There are five types of such anomalies: skyrocketing peak (SKP), plummeting peak (PLP), flat plateau (FPT), flat sink (FSK), and phantom peak (PP). Figure\,\ref{fig:image_label4} shows examples of such peak patterns from the FDOM time series data.  Skyrocketing peaks are characterized by a sharp upward spike or a narrow peak with a short base width, while a sharp downward spike characterizes plummeting peaks. These types of peaks may be caused by electronic sensor noise. Flat plateaus and flat sinks are characterized by a nearly constant signal amplitude at the top (plateau) and the bottom (sink), respectively, and may be caused by sediment deposits near or around the sensors. Flat sinks are only observed in FDOM data. Phantom peaks appear as normal peaks, but do not have a preceding stage rise that would trigger the peak. Non-hydrological events, such as animal activity in the water near the sensor may be the cause. To detect phantom peaks and plummeting peaks, it is necessary to consider the relationships between two data time series, while the other peak types can be identified using only one type of time series data.

\section{The AutoML Pipeline of \frameworkname{} Framework}
\label{sec:Pipeline}
 
\begin{figure}[ht]
\centering
\hspace*{-0.35in}\includegraphics[width=1.2\textwidth]{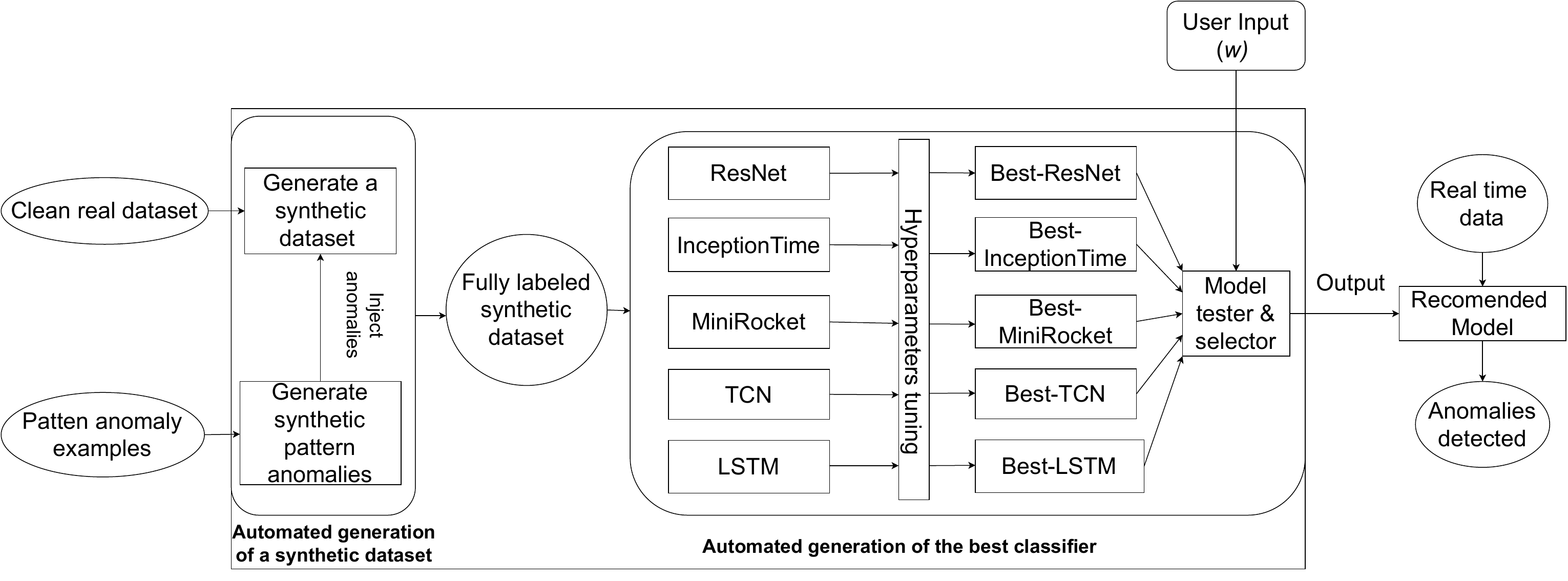}\\

\caption{The implemented \frameworkname{} automated supervised machine learning framework.}%architecture and processes.}
\label{fig:image_label5}
\end{figure}

The fully automated pipeline of \frameworkname{} framework is divided into two parts: one that automates creating a training set, and another that generates the best deep learning classifier through the tuning of architectural and training parameters of each model in the given pool. The generation of a model involves building and comparing different architectural instances of the model in conjunction with different training parameters. Figure\,\ref{fig:image_label5} shows an instance of the framework implemented in the current work. In this implementation, \frameworkname{} includes a range of sub-models drawn from a pool of state-of-the-art deep learning models, such as InceptionTime, MiniRocket, ResNet, TCN and LSTM as well as tools for generating time series data, injecting pattern anomalies into synthetic data, and tuning hyperparameters. 

\subsection{Synthetic data generation}

To generate synthetic watershed time series data (WTSD), we utilize the state-of-the-art time series generator TimeGAN, which uses a generative adversarial network (GAN) to output data that is nearly identical to the input data. We begin by obtaining a small portion of clean WTSD, such as clean data of one year, and use it to generate a large amount of synthetic data with TimeGAN. 

To create a labeled dataset for supervised learning, we augment and inject anomalies into synthetic data using a small number of ground truth labels. This process enables us to create a sufficiently large training dataset with minimal manual labeling while also addressing data sparsity and skewness issues that are common in watershed time series data.

\begin{figure}[ht]
\centering
\includegraphics[width=0.8\textwidth]{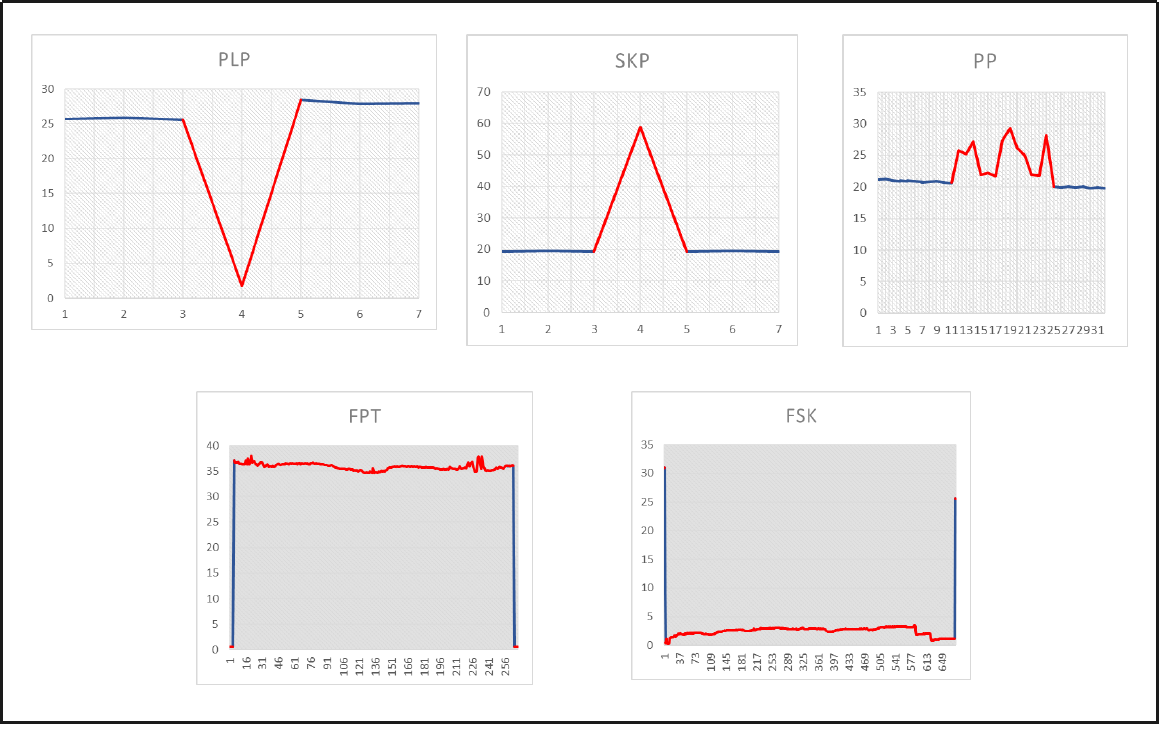}\\
\caption{Labeled anomalous peak patterns injected into synthetic time series data.}
\label{fig:anomalous_peak_patterns}
\end{figure}

To generate synthetic anomalies, we, again, utilize the state-of-the-art time series generator TimeGAN. By generating multiple altered versions of the identified peak pattern anomalies, we have obtained a sufficient number of instances of each anomaly type to train the deep learning models. These synthetic anomalies are then injected at random positions within the synthetic FDOM and turbidity time series data generated by TimeGAN to mimic the random occurrence of anomalies in real data. This results in a fully prepared and labeled training dataset for the deep learning classifiers. The importance of this step lies in creating a multi-class labeled peak pattern anomaly dataset suitable for training deep learning classifiers. Figure\,\ref{fig:anomalous_peak_patterns} shows the typical labeled peak-patterns anomalies injected into the generated synthetic time series data.

\subsection{Generating the best deep learning classifier}

\frameworkname{} handles the best model generation problem as an optimal search problem in a parameter space pertaining to the models. Each model has its own search space that includes a range of individual architectural and training hyperparameters to choose from. These hyperparameters are automatically tuned using optimizers to find the best model instance from a pool of select models. This automated process %makes the task of more deep-learning user-friendly 
is particularly helpful for hydrologists who may not have adequate expertise in machine learning. 

%\subsubsection{Model selection from a pool of deep learning classifiers}
\subsubsection{Model instance search using hyperparameter optimization}
\label{sec:hyperparm_opt}

Algorithm\,\ref{alg:autoML} outlines the AutoML algorithm of the \frameworkname{} framework. This algorithm tunes each model in the given model pool one at a time using hyperparameter optimization techniques and outputs a model instance expected to achieve the top performance based on the evaluation results. %By automating this process, we can identify the most suitable model efficiently and effectively, improving the accuracy and performance of our anomaly detection system.
\begin{algorithm}[ht]
\caption{AutoML algorithm of \frameworkname{} against the WTSD.}\label{alg:autoML}
%\SetKwFunction{isOddNumber}{isOddNumber}
\SetKwInOut{KwIn}{Input}
\SetKwInOut{KwOut}{Output}
\KwIn{a pool of models \{$M_1$, $M_2$, ..., $M_n$\};\\ synthetic watershed time series data (WTSD);\\ user's performance preference;}
\KwOut{the model instance showing the highest performance for the WTSD;}
%\tcc{Comment?.}
\For{{\rm each model} $M_i$ {\rm (}$i=1,2,...,n${\rm ) in the pool}}{
   %Build and 

   Generate the best model instances $\hat{m}_i$ from the models in the pool that achieves the highest accuracy during training on synthetic WTSD by tuning $M_i$'s architectural and training parameters through hyperparameter optimization;
   
   Get the user's performance preference $w$ and recommend the best model instance using Equation \ref{eqn:ImSC};
   
   Test the recommended best model instance $Tr(\hat{m}_i)$ against the real test dataset  to detect peak-pattern anomalies;
}
Return the trained model instance that has the highest performance score in the result pool;
\end{algorithm}

There are three aspects important to the efficacy of Algorithm\,\ref{alg:autoML}: search space, search strategy, and evaluation strategy. Each is discussed below.

The \emph{search space} is defined by a set of hyperparameters and their ranges. These ranges can be defined based on the specific needs and knowledge of the user. In our implementation of \frameworkname{}, the hyperparameters are the machine learning models in the input pool, the architectural parameters pertaining to each model (see Table\,\ref{tab:arc-hyperparam}), and the training parameters that are common across all models (see Table\,\ref{tab:train_hyp}). Overall, the search space allows for thoroughly exploring and optimizing various hyperparameters to identify the most suitable model instance and hyperparameter settings for a given data set. 

%%%%%%%%%%%%%%%%%%%%%%%%%%%%%%%%%%%%%%%%%%%%%%%%%%%
\begin{table}[ht]
\scriptsize    
    \begin{subtable}[ht]{0.45\textwidth}
        \centering
        \begin{tabular}{l | l }
        Hyperparameter & Domain \\
        \hline \hline
        Number of layers  & [18, 34, 50, 101, 152]\\
        Number of filters  & [16 -- 1024]\\
        Kernel size  & [1, 3, 5, 7]\\
        Stride  & [1, 2]\\
        Padding  & [0, 1]\\
        Pooling layer window size & [2$\times$2, 3$\times$3] 
       \end{tabular}
       \caption{ResNet.}
       \label{tab:ResNet_arch_param}
    \end{subtable}
\hfill
\scriptsize   
    \begin{subtable}[ht]{0.45\textwidth}
        \centering
        \begin{tabular}{l | l }
        Hyperparameter  & Domain \\
        \hline \hline
        Number of Inception modules & [1 -- 6]\\
        Number of filters & [32 -- 512]\\
        Filter size & [3, 5, 7, 11]\\
        Stride  & [1, 2]\\
        Pooling layer window size  & [3 -- 7]\\
        Dropout rate & [0.1 -- 0.5]
        \end{tabular}
        \caption{InceptionTime.}
        \label{tab:Inception_arch_param}
    \end{subtable}
    \hfill
    \medskip\\
\scriptsize
    \begin{subtable}[ht]{0.45\textwidth}
        \centering
        \begin{tabular}{l | l }
        Hyperparameter & Domain \\
        \hline \hline
        Number of random kernels  & [100 -- 5000]\\
        Kernel sizes  & [7 -- 21]\\
        Subsampling factor  & [2 -- 10]\\
        Normalization & [true, false]\\
        Number of random Fourier features  & [1000 -- 5000]
        \end{tabular}
       \caption{MiniRocket.}
       \label{tab:MiniRocket_arch_param}
    \end{subtable}
   \hfill
\scriptsize
    \begin{subtable}[ht]{0.45\textwidth}
        \centering
        \begin{tabular}{l | l }
        Hyperparameter & Domain \\
        \hline \hline
       Number of layers & [1 -- 5]\\
        Number of hidden units & [16 -- 512] \\
        Dropout rate & [0.1 -- 0.5]\\
        Recurrent dropout rate & [0.1 -- 0.5]\\
        Bidirectional & [yes, no]\\
        Activation function & [Sigmoid, Tanh, ReLU] \\
        Recurrent activation function & [Sigmoid, Tanh, ReLU]\\
        Layer normalization & [yes, no]   
        \end{tabular}
       \caption{LSTM.}
\label{tab:LSTM_arch_param}
    \end{subtable}
    \hfill
    \medskip
 \scriptsize   
\begin{subtable}[ht]{0.45\textwidth}
        \centering
        \begin{tabular}{l | l }
        Hyperparameter  & Domain \\
        \hline \hline
        Number of layers  & [2 -- 100]\\
        Kernel size & [1, 3, 5]\\
        Dropout rate & [0.1 -- 0.5]\\
        Number of input channels  & [1 -- 64]\\
        Number of filters  & [32 -- 1024]\\
        Stride & [1, 2]\\
        Dilation & [1 -- 4]\\
        Padding & [0, 1]
        \end{tabular}
        \caption{TCN.}
        \label{tab:TCN_arch_param}
     \end{subtable}
\caption{Architectural hyperparameters of the individual deep learning model types used
in HF-PPAD.}
     \label{tab:arc-hyperparam}
\end{table}
%%%%%%%%%%%%%%%%%%%%%%%%%%%%%%%%%%%%%%%%%%%%%%%%%%%%%%%%%%%%%%%%%%%%

\begin{table}[ht]
\scriptsize
\centering
\begin{tabular}{l | l }
Hyperparameter  & Domain\\
\hline \hline
Batch size & 32, 64, 128, 256, 512\\
Optimizer & SGD, Adam \\
Learning rate & 1e-6, 1e-5, 1e-4, 1e-3, 1e-2\\
Regularization & L1, L2, dropout\\
\end{tabular}
\caption{Training hyperparameters common to all the deep learning models in the pool.}
\label{tab:train_hyp}
\end{table}

%%%%%%%%%%%%%%%%%%%%%%%%%%%%%%%%%%%%%%%%%%%%%%%%%%%%%%%%%

The \emph{search strategy} determines the process for  iteratively selecting and evaluating combinations of hyperparameter values within the search space. The search strategy may be modified based on prior evaluations to improve future trials, or it may loop through all possible combinations within the search space. An effective search strategy can reduce the time required for the optimization process.
For this work, we use Optuna, a tool for hyperparameter optimization that includes the four hyperparameter optimizers chosen in this work (i.e., random forest, Bayesian, Hyperband, and greedy). These optimizers are included as hyperparameters themselves in the search space, and on each trial, the AutoML algorithm selects the optimizer that provides the best result. The select optimizer then optimizes the architectural and training hyperparameters of the chosen model. The search time is directly proportional to the number of trials conducted. % in the experiment. 
Increasing the number of trials can improve the results but can also increase the tuning time.

The \emph{evaluation strategy} is crucial, as it determines how the effectiveness of a model is evaluated with respect to its hyperparameters. The evaluation criteria, such as the validation performance and the total number of model parameters, are typically the same as those used in manual tuning. We also consider such factors as time/epoch, the number of parameters, and the memory usage for each model. By thoroughly evaluating the performance of each model and its corresponding hyperparameters, \frameworkname{} can identify the most suitable model instance for a given data set.

\subsubsection{User preference-based best model instance selection}

Our automated peak-pattern anomaly detection framework \frameworkname{} assists users in identifying the most appropriate machine learning model for their WTSD. The algorithm conducts exhaustive tuning of architectural and training hyperparameters to determine the optimal instance of each model. The effectiveness of each model instance is then evaluated using Equation\,\ref{eqn:ImSC}, which takes into account both the accuracy achieved and the computational cost incurred during the tuning process. The user is also asked to specify a weight indicating the relative importance of lower computational cost (e.g., training time and memory usage) compared to higher accuracy. This weight is linearly related to the computational cost and allows for personalized model instance recommendations based on the user's specific needs and preferences. By doing so, it helps users to make informed decisions about the best model instance for their specific data set, considering both performance and computational cost.

\begin{equation}
Q_{m_i} = (1-w) A_{m_i} + w(1-S_{m_i})\label{eqn:ImSC}
\end{equation}
\noindent where $m_i$ $(i=1,2,\dots,n)$ is an instance of a model $M_i$ in the pool; $Q_{m_i}$ is the output quality achieved using the model instance $m_i$; $A_{m_i}$ is the accuracy achieved using the model instance $m_i$; $S_{m_i}$ is the size of $m_i$ normalized by the maximum possible size of all instances of $M_i$ and $w$ is the user-provided weight of a smaller model size (i.e., $(1-S_{m_i})$) over higher accuracy (i.e., $A_{m_i}$). The size of a deep learning model can be determined by looking at the number of parameters. To determine the size of a deep learning model, we convert the total number of parameters to a more readable format, such as megabytes (MB) or gigabytes (GB), by dividing by the number of bytes per parameter (usually 4 for float32 data type).

\section{Experiments}
\label{sec:exp}
The \frameworkname{} implementation performed on the WTSD used here has been evaluated thoroughly. There are three main questions answered through experiments:
\begin{itemize}
    \item How similar is the synthetic time series dataset (with labeled peak-pattern anomalies injected) to the original real dataset from the WTSD?  (See Section\,\ref{sec:synthdataset}.)
    \item How well do the generated best individual deep learning models perform?  (See Section\,\ref{sec:ADperformance}.)
    \item How well does the autoML pipeline adapt to the user-specified preference between accuracy and computational cost to select the deep learning model that meets the preference best? (See Section\,\ref{sec:adaptpref}.)
\end{itemize}

\subsection{Setup}

%The experimental setup for this study involved the use of an AutoML framework, \frameworkname{}, to identify anomalous peak-pattern anomalies in WTSD. 

\paragraph{Datasets}
One year (from October 1, 2016 to September 30, 2017) worth of clean FDOM and turbidity data was used in all the experiments. %collected from domain scientists.
This dataset has 105,120 points.  They were passed to TimeGAN, the machine learning algorithm selected for generating synthetic labeled data, to create a dataset of 1,048,575 time series samples; the dataset was then split into 70\% training and 30\% validation datasets. TimeGAN was then trained for 5,000 epochs, as recommended by \cite{yoon_2019}, to ensure that it captured the main features and patterns of the real data.
Subsequent to the generation of the clean synthetic data, TimeGAN was used to generate synthetic instances of anomalous peak patterns (see Figure\,\ref{fig:anomalous_peak_patterns}).  Synthetic versions of 400 to 500 anomalous peak patterns were created for each type of anomaly and randomly injected into the synthetic dataset. The resulting dataset containing clean time series samples interspersed with anomalous peak patterns was used to train and validate the deep learning models in the pool. The trained models were then tested on real data containing real peak pattern anomalies.

\paragraph{Deep learning models}
The deep learning models in the pool included InceptionTime, MiniRocket,ResNet,TCN and LSTM. Each model has its own search space for architectural hyperparameters and a common search space for training hyperparameters as discussed in Section\,\ref{sec:hyperparm_opt}. The tuning of these hyperparameters was carried out using Optuna, a hyperparameter optimization library. Four such optimizers, including random forest, Bayesian, Hyperband, and greedy search, were included in the search space to find the best model instance for each deep learning model. The hyperparameter optimization process for each deep learning model was run for 1,000 trials with early stopping triggered when the validation loss did not improve for ten consecutive epochs. For validation, we used 70\% of the training dataset, selected through shuffling. The resulting best model instances of the models were then trained for 50 epochs and tested against the WTSD test dataset using the user-provided performance objective (see Equation\,\ref{eqn:ImSC}). The model instance that achieves the highest performance score in the test was then output.

\paragraph{Performance metrics}
For the anomaly detection task, the performance achieved by a trained deep learning model comprises accuracy and computational cost. The accuracy used in this work are balanced accuracy  (i.e., $\frac{1}{2} \left(\frac{TP}{TP + FN} + \frac{TN}{TN + FP}\right)$) and F-1 score (i.e., $2 \cdot \frac{Precision \cdot Recall}{Precision + Recall}$).  The computational costs are the time and memory consumed during model training. For simplicity, we use the number of model parameters as a proxy measure of computational cost, as both the training time and memory are proportional to it.  We also report other parameters relevant to the model training, such as validation loss, epoch time, and the number of epochs.

\paragraph{Computing platform}
All experiments were performed on Google Colab Pro platform, which provided access to a NVIDIA Tesla T4 GPU with 16GB of memory and an Intel Xeon E5-2670 v3 CPU with 8 cores and 30GB of memory. The programming language used was Python, with libraries including PyTorch and pandas.

\subsection{Similarity of the synthetic dataset to the real dataset}% using RNN regression model}
\label{sec:synthdataset}

\begin{figure}[ht]
\centering
\includegraphics[width=0.7\textwidth]{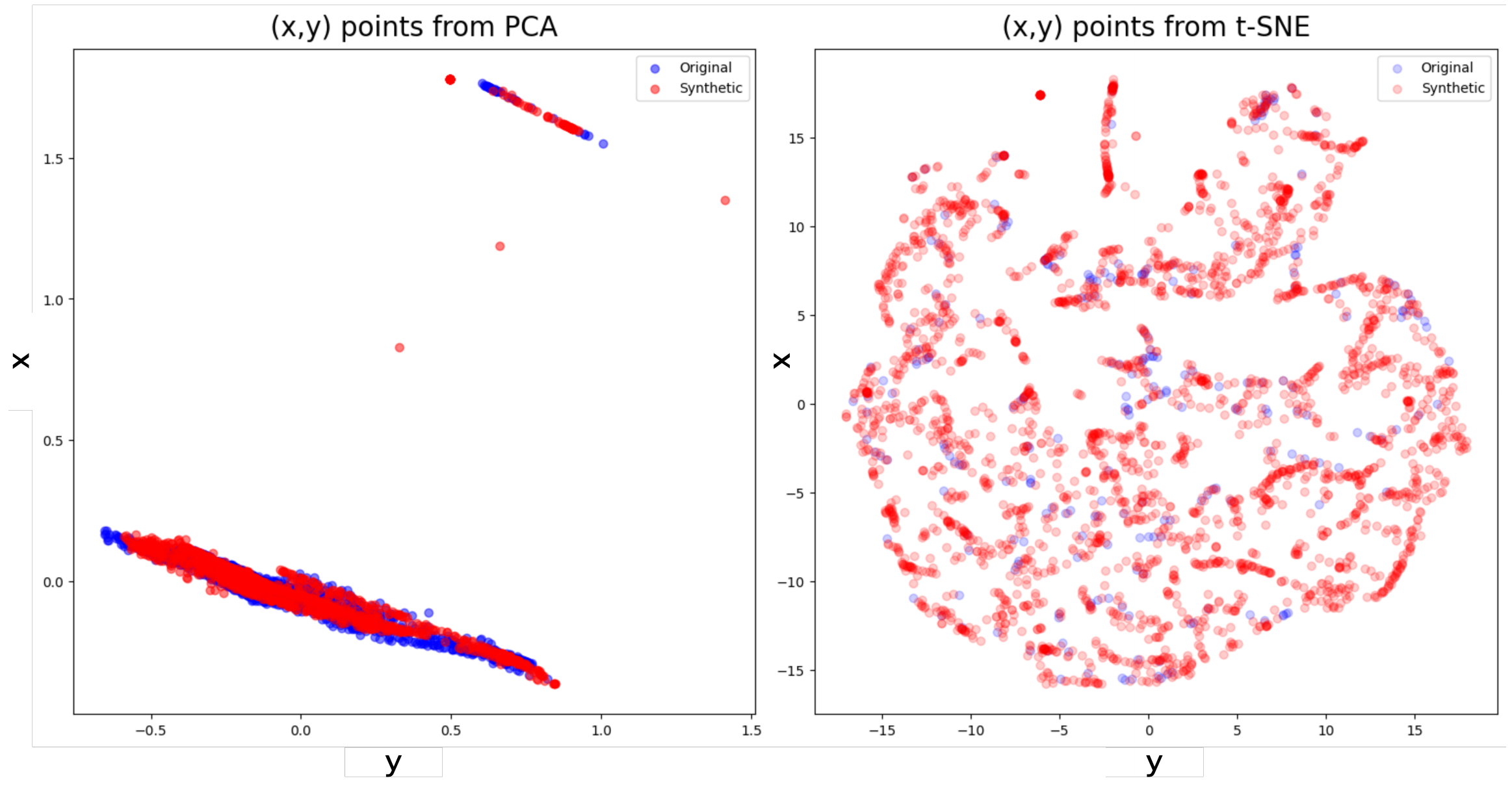}
\caption{Clusters of the synthetic and the original time series data points in a 2D turbidity $\times$ FDOM space generated by PCA (left) and t-SNE (right).} 
\label{fig:PCA_TSNE_plots}
\end{figure}

As mentioned, the synthetic data points were generated using TimeGAN based on a clean dataset collected from the WTSD at SRRW.
In order to evaluate the accuracy of the generated synthetic dataset, we selected two dominant variables, turbidity (for the x axis) and FDOM (for the y axis), from stage, turbidity, and FDOM through dimensionality reduction by PCA and by t-SNE, respectively, and generated clusters of the resulting data points in the 2D space of turbidity $\times$ FDOM. Figure\,\ref{fig:PCA_TSNE_plots} shows the clusters of data points generated through PCA (left) and t-SNE (right). In both plots, the clusters of the original data points (blue) and the synthetic data points (red) are almost the same, which demonstrates the high similarity between the real and synthetic datasets. 

To further verify the similarity, we trained an RNN regression model separately on real data and on synthetic data, and tested the two trained models against a separate real dataset. The RNN model consists of a single GRU (Gated Recurrent Unit) layer with 12 units and a dense output layer with six units and a Sigmoid activation function. The optimizer used is Adam and the loss function is mean absolute error (MAE). 
Table\,\ref{tbl:RNN_test} summarizes the test accuracy (R-squared ($R^2$), mean absolute error (MAE), and mean squared error (MSE)) achieved by the two trained RNN models. The test accuracy of the model trained on the synthetic data was close to the test accuracy of the model trained on real data (within 4\% for R$^2$, 2\% for MAE, and 5\% for MSE), confirming that the synthetic data generated by TimeGAN is a suitable substitute for real data in training the machine learning models for the WTSD.

\begin{table}
\centering
\begin{footnotesize}
\begin{tabular}{|l|l|l|l|} 
\hline
\multirow{2}{*}{\textbf{Training data}} & \multicolumn{3}{c|}{\textbf{Test accuracy}}\\
\cline{2-4}
 & \textbf{R$2$} & \textbf{MAE} & \textbf{MSE}\\ 
\hline
\textbf{Synthetic} & 0.301858  & 0.016981 & 0.003859\\
\hline
\textbf{Real} & 0.315577 & 0.016683 & 0.003672\\ 
\hline
\end{tabular}
\end{footnotesize}
\caption{Test accuracy of RNN regression models trained on synthetic dataset and real dataset and then tested on real dataset.}
\label{tbl:RNN_test}
\end{table}

\subsection{Anomaly detection performances of the best instances of the models}% trained on the synthetic dataset}
\label{sec:ADperformance}

\frameworkname{} generated best model instances and used synthetic datasets generated from the WTSD for training, and then tested the trained best model instances on the real dataset. Tables \ref{Table:2} and \ref{Table:3} summarize the performance results for each best trained model instance from the pool of models. All five models achieved high accuracy (70.2\% to 97.3\% for balanced accuracy and 64.7\%  to  94.6\% for F-1 score across FDOM and turbidity), which indirectly affirms the best model generation ability of \frameworkname{}.  The computational costs varied more significantly than accuracy depending on the model. Notably, the best trained LSTM model instance, which achieved the lowest accuracy, also incurred the lowest computational cost. This observation confirms the trade-off that leads to user-provided performance preference addressed below in Section\,\ref{sec:adaptpref}.

\begin{table}
\scriptsize
\centering
\hspace*{-0.65in}
\scalebox{1.2}{\resizebox{1\textwidth}{!}{
\begin{tabular}{|l|l|l|l|l|l|l|} 
\hline
\textbf{Model} & \textbf{Balanced accuracy} & \textbf{F-1 score} & \textbf{\# of parameters} & \textbf{Training time} & \textbf{Epoch time} & \textbf{\# of epochs}\\ \hline
InceptionTime & 97.3\% & 93.6\% & 1,817,888 & 350.5 sec & 7 sec & 50\\ 
\hline
ResNet & 95.3\% & 90.1\% & 8,130,502 & 550.2 sec & 11 sec & 50\\ 
\hline
MiniRocket & 93.4\% & 88.2\% & 89,974 & 150.6 sec & 3 sec & 50\\ 
\hline
LSTM & 70.2\% & 64.7\% & 17886 & 50.8 sec & 1 sec & 50 \\
\hline
TCN & 90.7\% & 85.9\% & 68,556 & 60.2 sec & 1.2 sec & 50 \\
\hline
\end{tabular}
}}
\caption{FDOM peak-pattern anomaly detection performance by the best trained model instance of each model in the \frameworkname{}'s model pool. }
%'s pool of models.}
\label{Table:2}
\end{table}

%%turbidity results%%%%
\begin{table}
\scriptsize
\centering
\hspace*{-0.65in}
\scalebox{1.2} {\resizebox{1\textwidth}{!}{
\begin{tabular}{|l|l|l|l|l|l|l|} 
\hline
\textbf{Model} & \textbf{Balanced accuracy} & \textbf{F-1 score} & \textbf{\# of parameters} & \textbf{Training time} & \textbf{Epoch time} & \textbf{\# of epochs}\\ \hline
InceptionTime & 95.3\% & 89.9\% & 4,082,884 & 467.2 sec & 9.34 sec & 50\\ 
\hline
ResNet & 98.3\% & 94.6\% & 11,921,636 & 721.5 sec & 14.43 sec & 50\\ 
\hline
MiniRocket & 91.6\% & 85.1\% & 118,974 & 349.7 sec & 6.94 sec & 50\\ 
\hline
LSTM & 74.2\% & 67.7\% & 23886 & 70.8 sec & 1.4 sec & 50 \\
\hline
TCN & 88.1\% & 81.9\% & 96,556 & 90.4 sec & 1.8 sec & 50 \\
\hline
\end{tabular}
}}
\caption{Turbidity peak-pattern anomaly detection performance by the best trained model instance of each model in the \frameworkname{}'s model pool.}
\label{Table:3}
\end{table}

\begin{figure}[ht!]
\centering 
\includegraphics[width=1\textwidth]{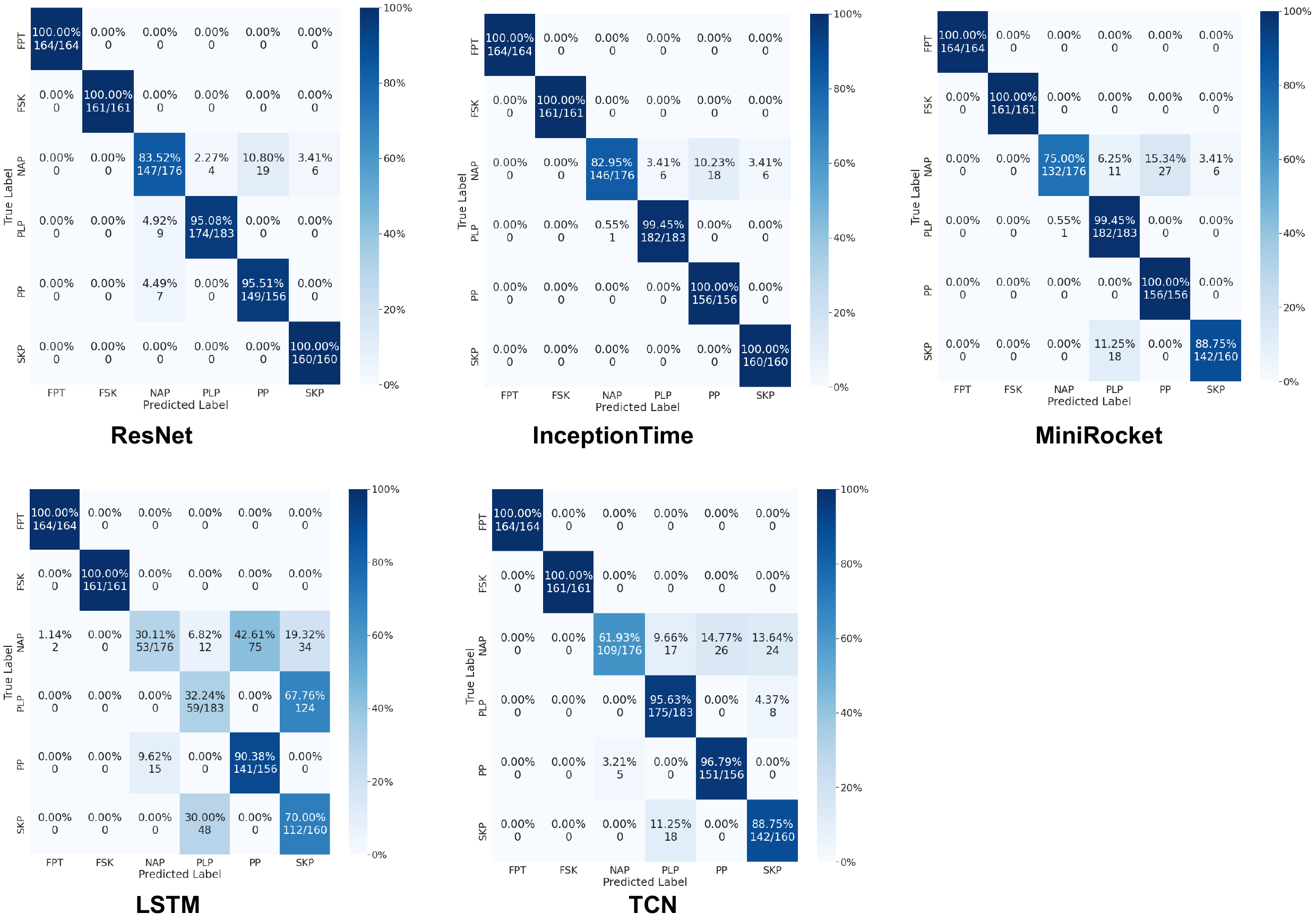}
\caption{Confusion matrix of FDOM peak-pattern anomaly detection accuracy by the best trained model instance of each model in the \frameworkname{}'s model pool. }
\label{fig:FDOM_CM}
\end{figure}

\begin{figure}[ht!]
\centering
\includegraphics[width=1\textwidth]{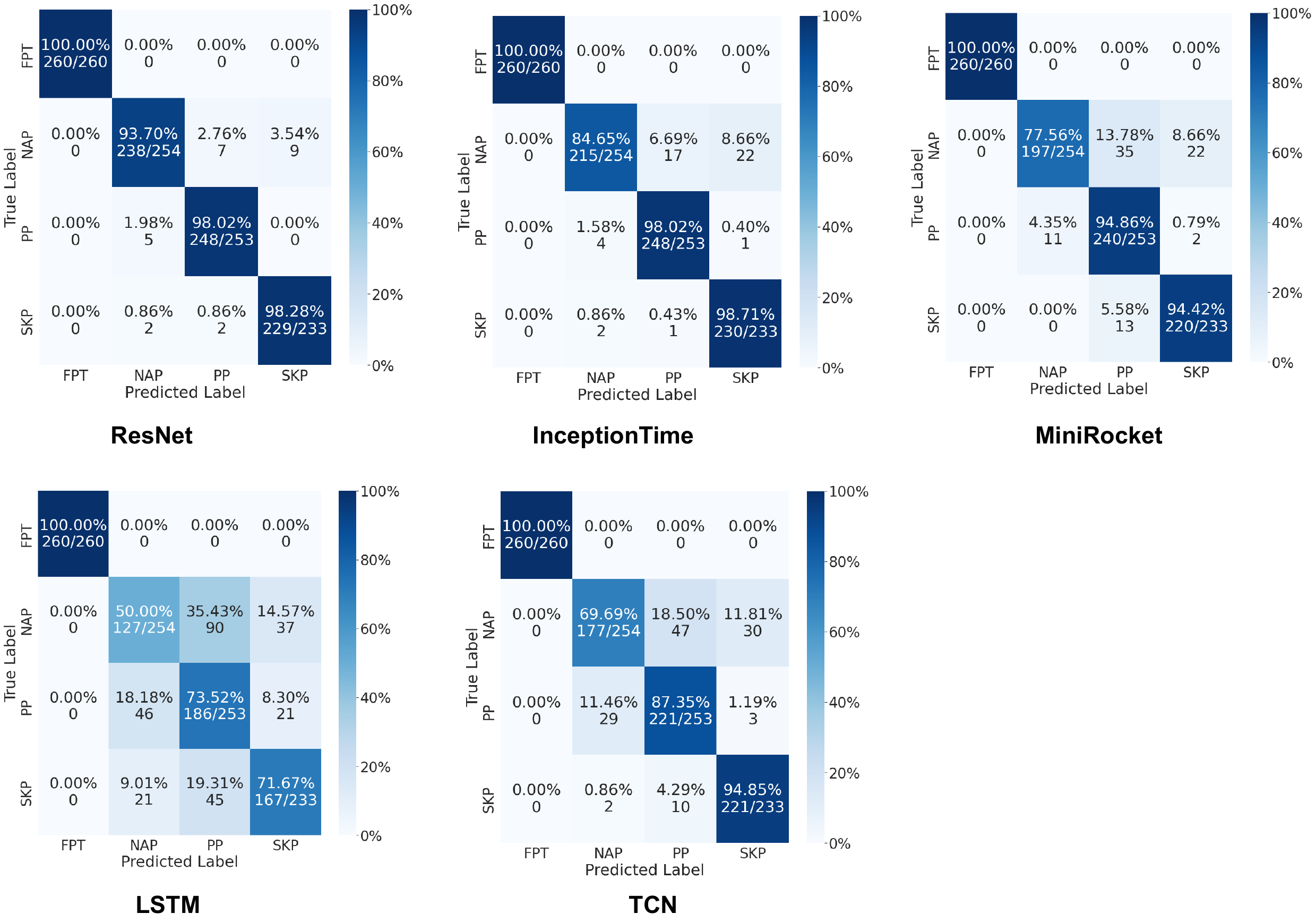}
\caption{Confusion matrix of turbidity peak-pattern anomaly detection accuracy by the best trained model instance of each model in the \frameworkname{}'s model pool.}
\label{fig:turbidity_CM}
\end{figure}

To further examine model performance with a focus on the anomaly detection accuracy, we have created the confusion matrices shown in Figure\,\ref{fig:FDOM_CM} for FDOM and Figure\,\ref{fig:turbidity_CM} for turbidity. Overall, the detection accuracy for all peak-pattern anomaly types are very high, which demonstrates the efficacy of the best model instance generation and training using the synthetic dataset. Particularly, the accuracy for the peak-pattern anomaly types FSK and FPT are 100\% for all the best model instances; we believe this accuracy is attributed to a long sequence of their anomaly instances that differentiate them from the other types of peak pattern anomalies. The accuracy for NAP is relatively lower than other anomaly types, as apparently some of them are mistaken as PP, PLP, or SKP peaks. Note that NAP is not an anomalous peak type. 

\subsection{User input based best model instance selection}% by \frameworkname{}}
\label{sec:adaptpref}
 
\begin{figure}[ht!]
\centering
\subfloat[For FDOM WTSD.]{\includegraphics[width=0.65\textwidth]{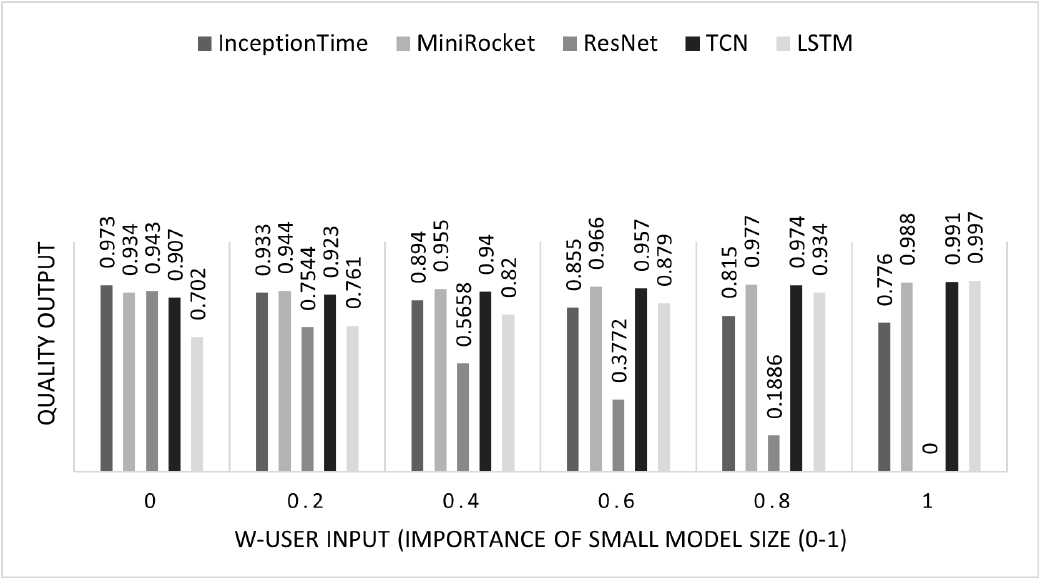}}
\medskip\\%\quad
\subfloat[For turbidity WTSD.]{\includegraphics[width=0.65\textwidth]{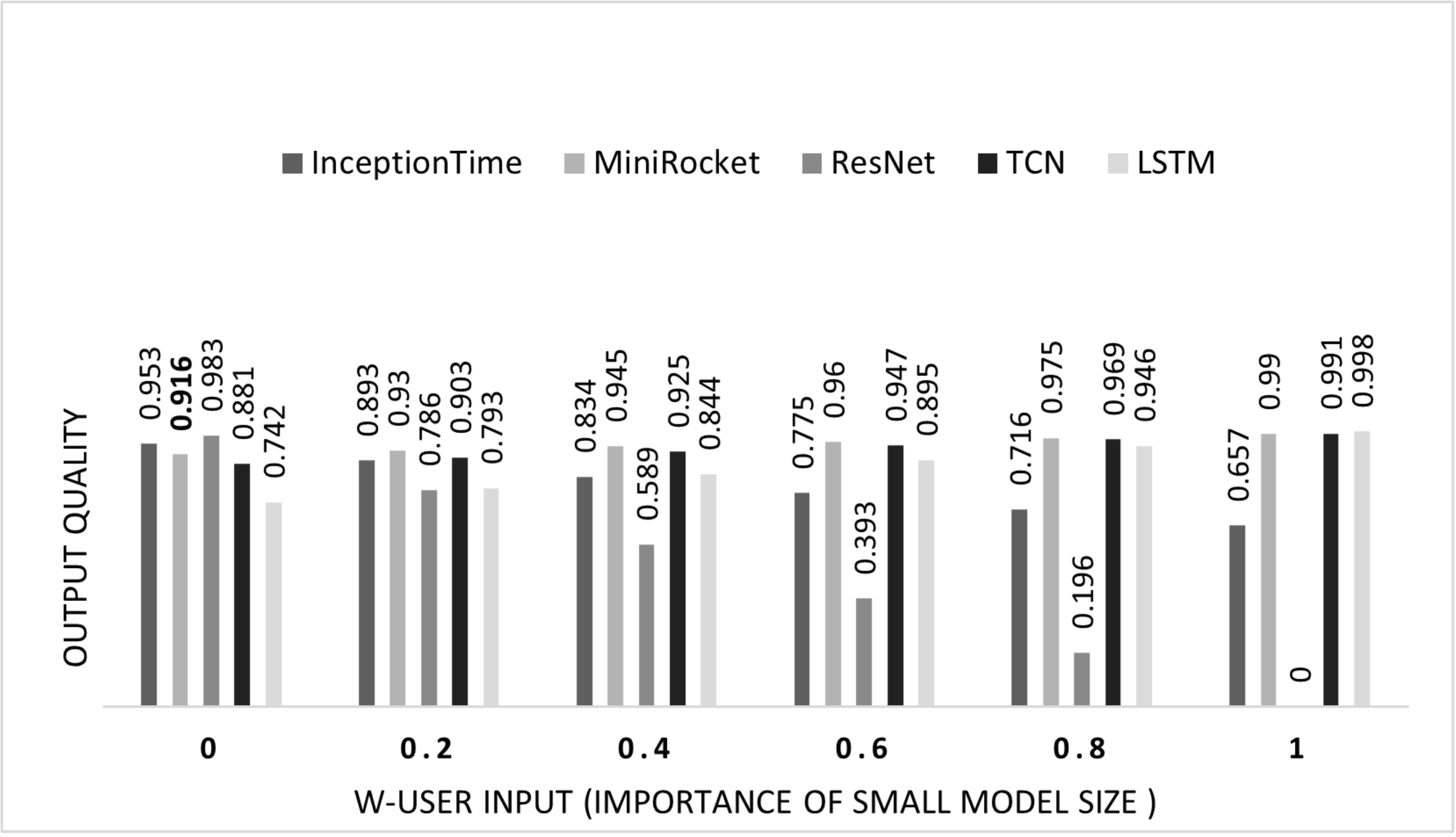}}
\caption{Comparison of the output quality achieved by the best model instance of each model for different values of the weight $w \in [0, 1]$; the weight indicates how much the user prefers small model size to high accuracy.}
\label{fig:quality_w}
\end{figure}

Recall that, the HF-PPAD approach recommends the best model instance for a dataset based on user preferences for accuracy and model size. Output quality was measured for the best trained model instance of each model using Equation\,\ref{eqn:ImSC}) and varying the weight parameter $w$ from $0$ to $1$ at the increment of $0.2$ for the FDOM and turbidity datasets. The results are shown as clustered bar charts in Figure\,\ref{fig:quality_w}. The InceptionTime model instance had the highest accuracy for FDOM (0.973) at $w=0$, whereas the TCN and MiniRocket model instances achieved the highest output quality (0.977 and 0.974, respectively) at $w=0.8$. For turbidity, ResNet had the highest accuracy (0.983) at $w=0$, while TCN and MiniRocket had the highest output quality (0.975 and 0.969, respectively) at $w=0.8$. We can summarize that TCN and MiniRocket are recommended for users who prioritize accuracy and low computational cost, while InceptionTime and ResNet are best for users who prioritize high accuracy; and additionally that LSTM is recommended for users who prioritize low computational cost, despite its lower accuracy, as it has a smaller model size compared to the other models.

%line Marked_graph

\begin{figure}[ht!]
\centering
\subfloat[For FDOM WTSD.]{\includegraphics[width=0.65\textwidth]{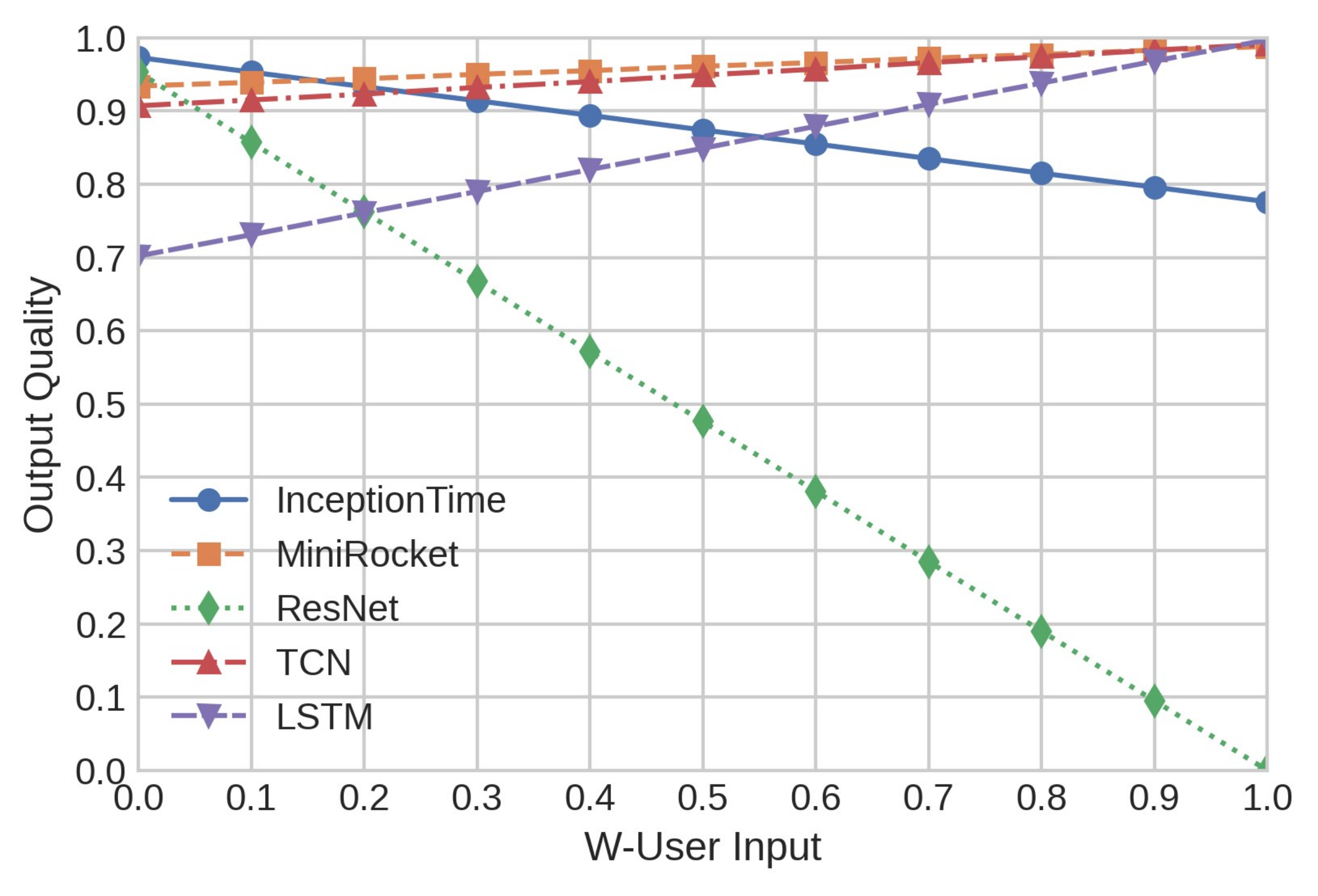}}
\medskip\\%\quad
\subfloat[For turbidity WTSD.]{\includegraphics[width=0.65\textwidth]{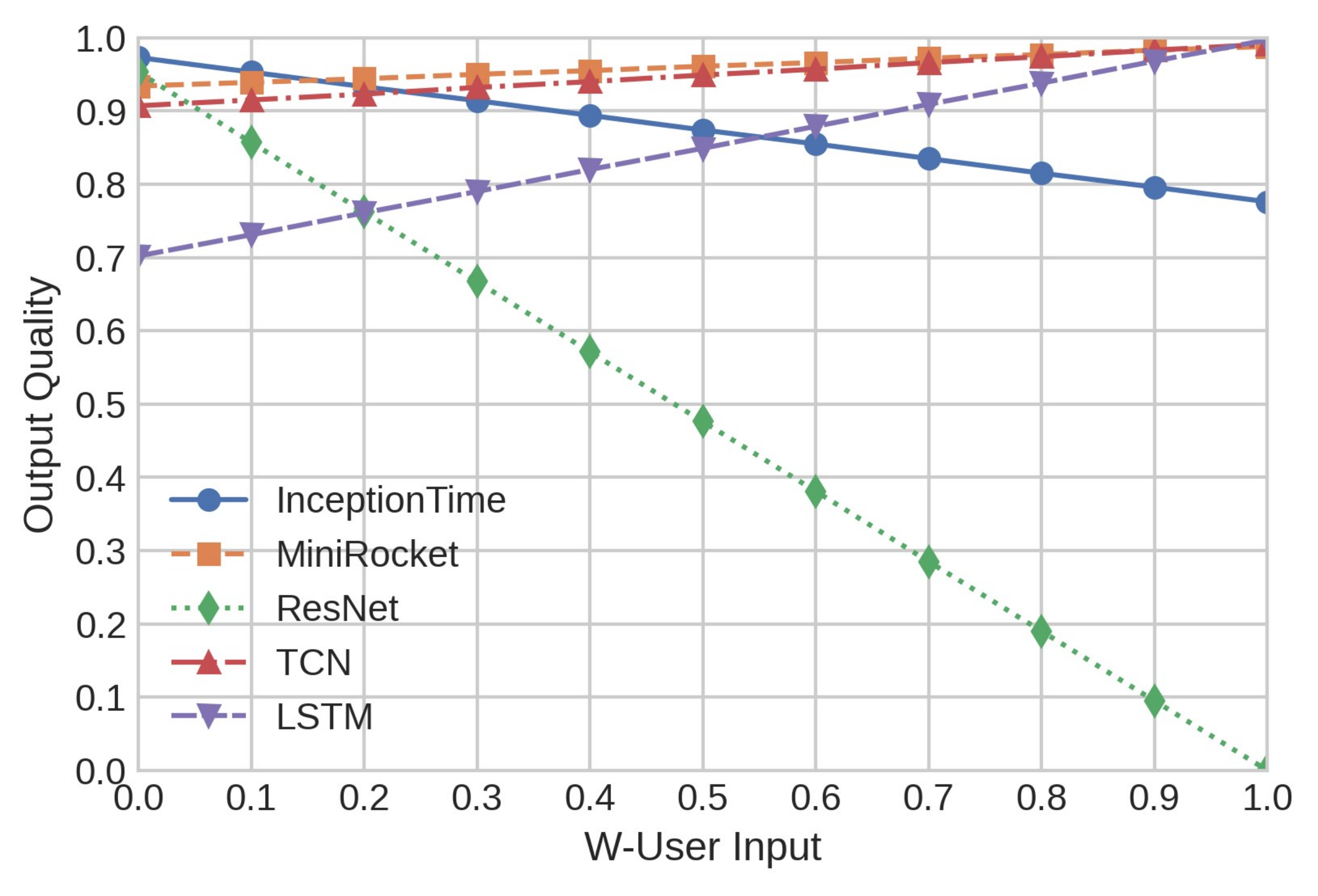}}

\caption{Changes of the output quality achieved by the best model instance of each model for the weight $w$ increasing from $0$ to $1$.}
\label{fig:Line_quality_w}
\end{figure}

 Figure\,\ref{fig:Line_quality_w} shows a line graph of the output quality of the best model instance of each model as the user preference input $w$ increases (at the increment of 0.1). It visualizes the trends of the output qualities changing between the different models. Specifically, it exhibits a decrease in the output quality of a model with a larger size as the $w$ value increases. Notably, the MiniRocket and TCN models are competitive options for users who prioritize accuracy and low cost computational requirements. In contrast, the LSTM model only achieves higher output quality when $w$ is 0 due to its smaller size. Overall, the figure highlights the varying output quality of the models and provides valuable insights into selecting the appropriate model based on user preferences.

%%%%%%%%%%%%%
 
\section{Conclusion}% and Future Work}
\label{sec:conc}
This paper presented an anomaly detection framework using automated machine learning (AutoML) on WTSD from the northeast US critical zone. The framework is designed to assist hydrologists in identifying anomalous events in their data, such as peak-pattern anomalies in FDOM and turbidity, without needing expert knowledge in machine learning or anomaly detection algorithms. The framework consists of two main components: a synthetic labeled dataset generator and an automated best model instance generator. During implementation, we used TimeGAN for the synthetic dataset generation, and used InceptionTime, ResNet, MiniRocket, TCN and LSTM as the models in the pool; then the model instance that is best overall considering both accuracy and computational cost (i.e. model size) was identified (for recommendation) according to the user preference.
 
Our work is the first to utilize automated machine learning for peak pattern anomaly detection in WTSD. Our approach includes synthetically generated time series data and thorough hyperparameter optimization for model generation, demonstrating the potential of AutoML for time series classification tasks in hydrology. Experiments conducted demonstrate the high performance achieved by our framework applied to WTSD. Our contribution offers an innovative approach for efficient peak pattern anomaly detection in WTSD, providing a valuable tool for hydrologists and related stakeholders in water management.

For future work, we plan to improve the framework by incorporating additional machine learning models and expanding the search space for model generation; we also plan to test the framework on a wider range of WTSD and other environmental sensors data (e.g., snow and air humidity) to validate its generalizability. Additionally, we plan to investigate the use of the framework for other domains of anomalous events, such as those observed in water quality monitoring and flood forecasting. By continuing to refine and expand the capabilities of the framework, we hope to make it an essential tool for hydrologists in their efforts to monitor and understand water resources.

\section{Acknowledgments}

This material is based upon work supported by the National Science Foundation under Grant No. EAR 2012123. Any opinions, findings, and conclusions or recommendations expressed in this material are those of the author(s) and do not necessarily reflect the views of the National Science Foundation. Any use of trade, firm, or product names is for descriptive purposes only and does not imply endorsement by the U.S. Government.
The work was also supported by the University of Vermont College of Engineering and Mathematical Sciences through the REU program.
The authors would like to thank the US Geological Survey (USGS) for offering the domain expertise that was crucial to identify the peak anomaly types that are of practical
importance.





\end{document}